%% file: main.tex
\documentclass{article}
\pdfoutput=1
\usepackage{arxiv}
\usepackage[utf8]{inputenc} 
\usepackage[T1]{fontenc}    
\usepackage{hyperref}       
\usepackage{url}            
\usepackage{booktabs}       
\usepackage{amsfonts}       
\usepackage{nicefrac}       
\usepackage{microtype}      
\usepackage{fancyhdr}

\usepackage{algorithmic}
\usepackage{algorithm}
\usepackage{multirow}
\usepackage{amssymb}
\usepackage{graphicx}
\usepackage{subcaption}

\usepackage{comment}

\usepackage{hyperref}
\usepackage{enumitem}

\input{math_commands}

\usepackage{todonotes}
\newtheorem{definition}{Definition}

\title{Meta-Solver for Neural Ordinary Differential Equations}

\author{
Julia Gusak\thanks{Corresponding author}\\
  Skolkovo Institute of Science and Technology\\
  Moscow, Russia \\
  \texttt{y.gusak@skoltech.ru}
   \And
 Alexandr Katrutsa \\
  Skolkovo Institute of Science and Technology\\
  Moscow, Russia \\
  \texttt{aleksandr.katrutsa@phystech.edu}
  \AND
  Talgat Daulbaev \\
  Skolkovo Institute of Science and Technology\\
  Moscow, Russia \\
  \texttt{talgat.daulbaev@skoltech.ru} \\
  \And
  Andrzej Cichocki\\
  Skolkovo Institute of Science and Technology\\
  Moscow, Russia \\
  \texttt{a.cichocki@skoltech.ru}
  \And
  Ivan Oseledets\\
  Skolkovo Institute of Science and Technology\\
  Moscow, Russia \\
  \texttt{i.oseledets@skoltech.ru}
}

\begin{document}

\maketitle

\input{1_abstract}
\input{2_introduction}
\input{3_rk_parametrization}
\input{4_meta_node}
\input{5_experiments}
\input{6_conclusion}

\bibliographystyle{unsrt}
\bibliography{lib}

\appendix

\input{7_appendix}

\end{document}

%% file: math_commands.tex
\usepackage{amsmath,amsfonts,bm}

\def\eqref#1{equation~\ref{#1}}

\def\1{\bm{1}}

\def\vtheta{{\bm{\theta}}}

\def\vz{{\bm{z}}}

\DeclareMathAlphabet{\mathsfit}{\encodingdefault}{\sfdefault}{m}{sl}
\SetMathAlphabet{\mathsfit}{bold}{\encodingdefault}{\sfdefault}{bx}{n}



%% file: 1_abstract.tex
\begin{abstract}
A conventional approach to train neural ordinary differential equations (ODEs) is to fix an ODE solver and then learn the neural network's weights to optimize a target loss function. However, such an approach is tailored for a specific discretization method and its properties, which may not be optimal for the selected application and yield the overfitting to the given solver. In our paper, we investigate how the variability in solvers' space can improve neural ODEs performance. We consider a family of Runge-Kutta methods that are parameterized by no more than two scalar variables. Based on the solvers' properties, we propose an approach to decrease neural ODEs overfitting to the pre-defined solver, along with a criterion to evaluate such behaviour. Moreover, we show that the right choice of solver parameterization can significantly affect neural ODEs models in terms of robustness to adversarial attacks. Recently it was shown that neural ODEs demonstrate superiority over conventional CNNs in terms of robustness. Our work demonstrates that the model robustness can be further improved by optimizing solver choice for a given task. The source code to reproduce our experiments is available at \url{https://github.com/juliagusak/neural-ode-metasolver}.
\end{abstract}

%% file: 2_introduction.tex
\section{Introduction}

Neural ODE models proposed by~\cite{chen2018neural} attract the  attention of the machine learning community since they demonstrate promising results in many application such as density estimation~\cite{grathwohl2018ffjord}, robustness to adversarial attacks~\cite{hanshu2019robustness,carrara2019robustness} and some others~\cite{chen2020mri,rubanova2019latent,de2019gru,giannone2020real,yang2019pointflow}. 
The main feature of this architecture is the combination of deep learning techniques and the theory of ODE systems.
This synthesis provides methods to derive interpretable architectures of neural networks and tools for their analysis.


The considered neural ODE models consist of standard neural network modules, e.g., convolutional layers, ResNet blocks, and so-called ODE blocks.
The forward pass through the ODE block is performed by numerical integration of the following initial value problem:
\begin{equation}
\begin{cases}
\frac{\mathrm{d} \vz}{\mathrm{d} t} = f(t, \vz(t), \vtheta), \quad t \in [t_0, t_1] \\ 
\vz(t_0) = \vz_0,  
\label{eq:fwd_1}
\end{cases}
\end{equation}
where $\vz_0$ is the output activations from the previous layer, $\vz(t_1)$ is the output of ODE block and the right-hand side $f$ is a parametric function that is trained by the backpropagation technique.
There are two approaches to perform backpropagation through the ODE block named Discretize-Optimize and Optimize-Discretize~\cite{onken2020discretize}.
Both approaches depend on the used ODE solver, but as far as we know, no papers explore how the choice of a numerical integration scheme affects the quality of the trained a neural ODE model.
In our paper, we aim to investigate this influence and show how the model performance can be improved by the choice of ODE solver.
Moreover, we consider a parametric family of ODE solvers and propose a method of exploiting this family to improve the quality of neural ODE models.


The typical choice of ODE solver in neural ODEs training is some ODE solver from the class of explicit Runge-Kutta methods with fixed or adaptive step size~\cite{dormand1980family}.
Further, we consider only explicit Runge-Kutta methods.
A Runge-Kutta method constructs the function approximation based on the following rule
\begin{equation}
    \begin{split}
    &\hat{\vz}_{k+1} = \hat{\vz}_k + h \sum_{i=1}^s b_i k_i, \quad k_i = f\left( t_k + c_i h, \hat{\vz}_k + h\sum_{j=1}^{s-1} w_{ij}k_j, \vtheta  \right),
    \end{split}
    \label{eq::rk_general}
\end{equation}
where $h$ is a step size, $\hat{\vz}_k$ is approximation of the ground-truth dynamic $\vz(t)$ in the grid point $t_k$.
The coefficients $c_i, b_i$ and $w_{ij}$ define a particular Runge-Kutta method and form the so-called Butcher tableau~\cite{butcher2008numerical}:
\begin{figure}[!ht]
\centering
\begin{tabular}{ c|ccccc } 
$0$ & $0$ & &  &  &  \\ 
$c_2$ & $w_{21}$ & $0$ &  &  &  \\ 
$c_3$ & $w_{31}$ & $w_{32}$ & $0$ &  &   \\ 
\dots & \dots & \dots & \dots &  &  \\ 
$c_s$ & $w_{s1}$ & $w_{s2}$ & \dots &  $w_{s, s-1}$ & $0$ \\ 
\hline
 & $b_1$ & $b_2$ & \dots &  $b_{s-1}$ & $b_{s}$ \\ 
\end{tabular}
\caption{The general form of Butcher tableau.}
\label{tab::butcher_tableau}
\end{figure}

The typical additional condition is $c_i = \sum_{j=1}^{i-1} w_{ij}$ for $i > 1$ and the consistency requirement leads to equality condition $\sum_{i=1}^s b_i = 1$. 
Also denote by $s$ a number of points in segment $[t_k, t_{k+1}]$, where the values of $f$ are computed.
Another name for this number is a \emph{number of stages} in the Runge-Kutta method.
One more important property of any Runge-Kutta method is the \emph{order of function approximation} denoted by $p$.
\begin{definition}
A Runge-Kutta method is of the order $p$ if the following inequality holds for any $\tilde{t} \in [t_0, t_1]$ such that $\tilde{t} + h \in [t_0, t_1]$: $\|\vz(\tilde{t} + h) -  \hat{\vz}(\tilde{t} + h)\| \leq Ch^{p+1}$, where $\vz(t)$ and $\hat{\vz}(t)$ are ground-truth and approximate dynamics.
\label{def::order}
\end{definition}
Runge-Kutta methods such that the number of stages $s$ equals to the order $p$, are of particular interest since the corresponding Butcher tableaux can be parametrized with no more than two scalar parameters~\cite{wanner1996solving}.
In our work, we consider the parametrizations of Runge-Kutta methods as one more degree of freedom to make neural ODE models better.



The paper~\cite{hanshu2019robustness} has shown that neural ODE models are more robust to adversarial attacks than convolutional neural networks (CNNs).
However, the authors ignore the dependence of the robustness on the used ODE solver in the forward and backward passes. 
We fill this gap and empirically demonstrate the influence of the ODE solver on neural ODE model performance in terms of both test accuracy and robustness to adversarial attacks.
Moreover, to adjust the ODE solver for performance improvement, we introduce the solver smoothing technique.
This technique aims to make a neural ODE model more robust to adversarial attacks.
In addition, this view on the neural ODE models and the role of ODE solver is related to meta-learning approach to train deep learning models~\cite{hospedales2020meta} since the choice of ODE solver affects the training of neural ODE model similar to meta-model affects the corresponding learner.     

Our main contribution is summarised as follows:
\begin{itemize}
    \item We empirically demonstrate that the choice of ODE solver significantly affects neural ODE model performance, particularly the robustness to adversarial attacks.
    \item We propose ODE solver smoothing of Runge-Kutta methods to improve the robustness of neural ODE models.
    \item We demonstrate that the proposed technique can be successfully combined with artificially nosing of data to additionally increase robustness of the trained model.   
\end{itemize}

\subsection{Related works}

The neural ODE research lies in the intersection of deep learning and the theory of ODEs, and thus takes an inspiration from both of them.
The modelling of dynamical systems with neural networks is discussed in many papers~\cite{greydanus2019hamiltonian,raissi2018multistep,ruthotto2019deep,chang2018multi}. 
One of the approaches is to describe a dynamical system as a solution of ODE given an initial value~\cite{brin2002introduction}.
Neural ODEs can be used to approximate such dynamics based on data samples.
In particular, models for time series prediction problem~\cite{rubanova2019latent,de2019gru} and data generation~\cite{grathwohl2018ffjord} can exploit neural ODEs architecture.   



Also, different discretization schemes of ODEs inspire researchers to build new deep learning architectures~\cite{lu2018beyond}.
In particular, PolyNet~\cite{zhang2017polynet}, RevNet~\cite{gomez2017reversible} and FractalNet~\cite{larsson2016fractalnet} are motivated by the backward Euler scheme, forward Euler scheme and the classical Runge-Kutta method, respectively.
Also, the single-image super-resolution problem is solved with neural network motivated by ODE integration scheme in~\cite{he2019ode}.


Another group of papers addresses issues of neural ODE training and discusses approaches to achieve competitive results compared with other architectures.
For instance, \cite{gholami2019anode},~\cite{zhuang2020adaptive} address the instability of the adjoint method with a checkpointing strategy, and \cite{daulbaev2020interpolated} proposes to use the interpolation technique in the backward pass.
The importance of the augmentation technique in the context of training neural ODE is presented in~\cite{dupont2019augmented}. 
The well-known fact from numerical analysis~\cite{wanner1996solving} is that to solve ODE with sufficient accuracy, a small step size in an ODE solver is required.
However, this setting leads to an increase of the running time to perform the forward pass in neural ODE.
This issue is addressed in~\cite{kelly2020learning,ghosh2020steer}, where the trained dynamic is forced to be easy to solve with regularization of the loss function and sampling of the integration final time, respectively.
Also, the extension of neural ODEs to stochastic neural ODEs is considered in~\cite{li2020scalable,liu2019neural,tzen2019neural,oganesyan2020stochasticity}.

Besides the standard machine learning quality measures, neural ODEs can be evaluated based on the properties of the learned dynamic.
The verification of the learned dynamic stability with respect to decreasing step size in the used ODE solver is studied in~\cite{gusak2020towards} where $(\mathcal{S}, n)$-criterion is proposed for that purpose. 
Further, similar analysis of the learned dynamic is performed in~\cite{ott2020neural}.
The related question on the importance of control trained dynamic properties is discussed in~\cite{queiruga2020continuous}, where the continuous-in-depth extension of ResNet architecture is proposed.

One of the crucial factors in the evaluation of machine learning models is robustness to adversarial attacks~\cite{akhtar2018threat}.
The studies~\cite{hanshu2019robustness,carrara2019robustness} demonstrate that neural ODEs are more robust to adversarial attacks than classical CNN models.
However, the dependence of the robustness on the used ODE solver is lacking in their research.
In our study, we fill this gap and investigate how the choice of ODE solver during training affects the resulting robustness of neural ODE model.

%% file: 3_rk_parametrization.tex
\section{Parametrizations of Runge-Kutta methods}

\textbf{Key idea:} Runge-Kutta methods typically used in the neural ODE training can be parametrized with no more than two scalar variables.

In this study, we consider explicit Runge-Kutta methods such that their order $p$ equals to the number of stages $s$.
This requirement leads to the constraints on the coefficients from Butcher tableau.
These constraints induce parametrizations of the Runge-Kutta methods that we will use to improve the performance of the neural ODE model. 
We provide the considered parametrizations of Runge-Kutta methods below following~\cite{wanner1996solving}.

\begin{figure}[!h]
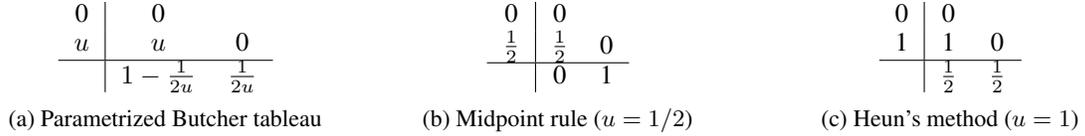

\centering
\begin{subfigure}{0.3\textwidth}
\centering
\begin{tabular}{ c|cc } 
0 & 0 &  \\ 
$u$ & $u$ & 0  \\ 
\hline
 & $1 - \frac{1}{2 u}$ & $\frac{1}{2 u}$
\end{tabular}
\subcaption{Parametrized Butcher tableau}
\label{tab::sp2}
\end{subfigure}
~
\begin{subfigure}{0.3\textwidth}
\centering
\begin{tabular}{ c|cc } 
0 & 0 &  \\ 
$\frac{1}{2}$ & $\frac{1}{2}$ & 0  \\ 
\hline
 & 0 & 1
\end{tabular}
\subcaption{Midpoint rule ($u = 1/2$)}
\label{tab::midpoint}
\end{subfigure}
~
\begin{subfigure}{0.3\textwidth}
\centering
\begin{tabular}{ c|cc } 
0 & 0 &  \\ 
1 & 1 & 0  \\ 
\hline
 & $\frac{1}{2}$ & $\frac{1}{2}$
\end{tabular}
\subcaption{Heun's method ($u=1$)}
\label{tab::heun}
\end{subfigure}
\caption{Examples of Butcher tableaux corresponding to 2-stage RK methods of the 2-nd order.}
\label{tab::rk2_butcher}
\end{figure}

\paragraph{Runge-Kutta methods of the 2-nd order with two stages.}
These Runge-Kutta methods are defined by the Butcher tableau whose coefficients have to satisfy the following system of equations:
\[
\begin{cases}
b_1 + b_2= 1\\
b_2c_2 = \frac{1}{2}.
\end{cases}
\]
Thus, these methods can be parametrized by a single parameter $u \in (0, 1]$; see the corresponding Butcher tableau in Figure~\ref{tab::sp2}.
\begin{definition}
Let parameters of Runge-Kutta methods be a set of values that uniquely define the Butcher tableau corresponding to the considered class of Runge-Kutta methods.
\end{definition}
In particular, midpoint rule and Heun's method are particular cases of such parametrization if $u = \frac{1}{2}$ and $u = 1$, respectively, see Figures~\ref{tab::midpoint} and~\ref{tab::heun}. 

Thus, adjusting parameter $u$ can improve performance of neural ODE models.
Techniques to adjust parameters of Runge-Kutta methods are presented in Section~\ref{sec::approach}.

Therefore, the natural approach to adjust ODE solver of the selected order during training is to vary the corresponding parameters, generate Butcher tableaux and exploit the corresponding ODE solver.
That is, we introduce an additional degree of freedom to the neural ODE setting. The motivation for that is provided in the next section.

\subsection{The choice of ODE solver affects robust accuracy of neural ODE model}

\paragraph{MNIST classification task.}
We illustrate the dependency of the robust accuracy on the choice of ODE solver on the MNIST classification task.
We consider the 2-nd order two stages Runge-Kutta methods for various values of parameter $u$ from the interval $(0,1]$.
The corresponding Butcher tableau is given in Figure~\ref{tab::sp2}.
We train 
ResNet-like model and
evaluate the robust accuracy using PGD attack with $\varepsilon = 0.3$, learning rate $2 / 255$ and 7 iterations. 
We provide the dependency of the robust accuracy on the value of solver parameter $u$ in Figure~\ref{fig:mnist_attack_demo}. 
Details on the architecture and training procedure can be found in Appendix. 

Also, in Figure ~\ref{fig:mnist_u} you can see the robust accuracy curve for three different values of $u$ for RK2 solver.
This plot illustrates that the different $u$ leads to different robust accuracy and this dependence is stable over the training epochs.
Here we use PGD attack with the same hyper-parameters.

\begin{figure}[!ht]
    \centering
    \begin{subfigure}[t]{0.49\textwidth}
    \centering
    \includegraphics[width=\linewidth]{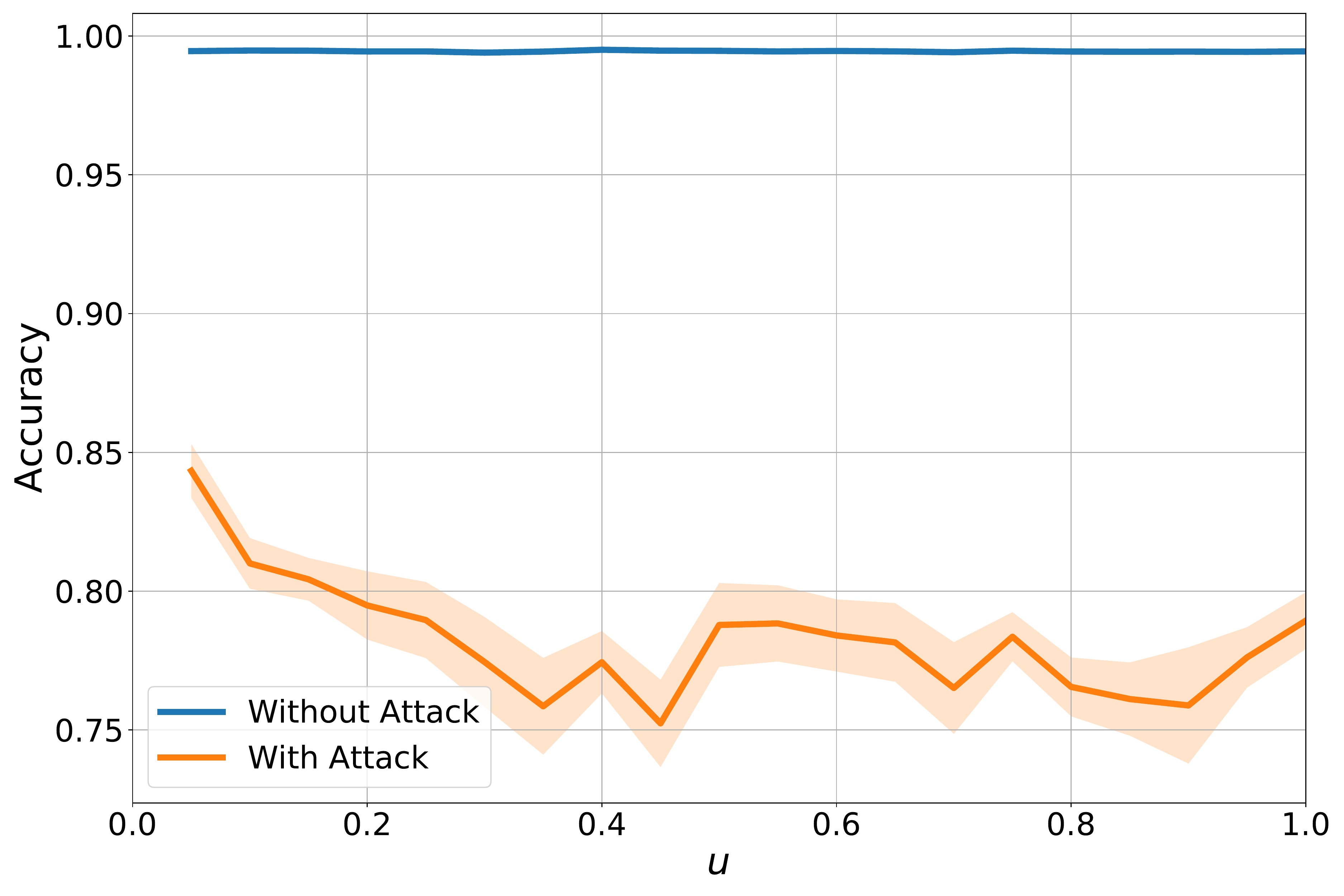}
    \caption{Robust accuracy of the model on MNIST dataset vs. different values of parameter $u$ in the 2-nd order Runge-Kutta solver (see Figure~\ref{tab::sp2}).}
    \label{fig:mnist_attack_demo}
    \end{subfigure}
    ~
    \begin{subfigure}[t]{0.49\textwidth}
    \includegraphics[width=\linewidth]{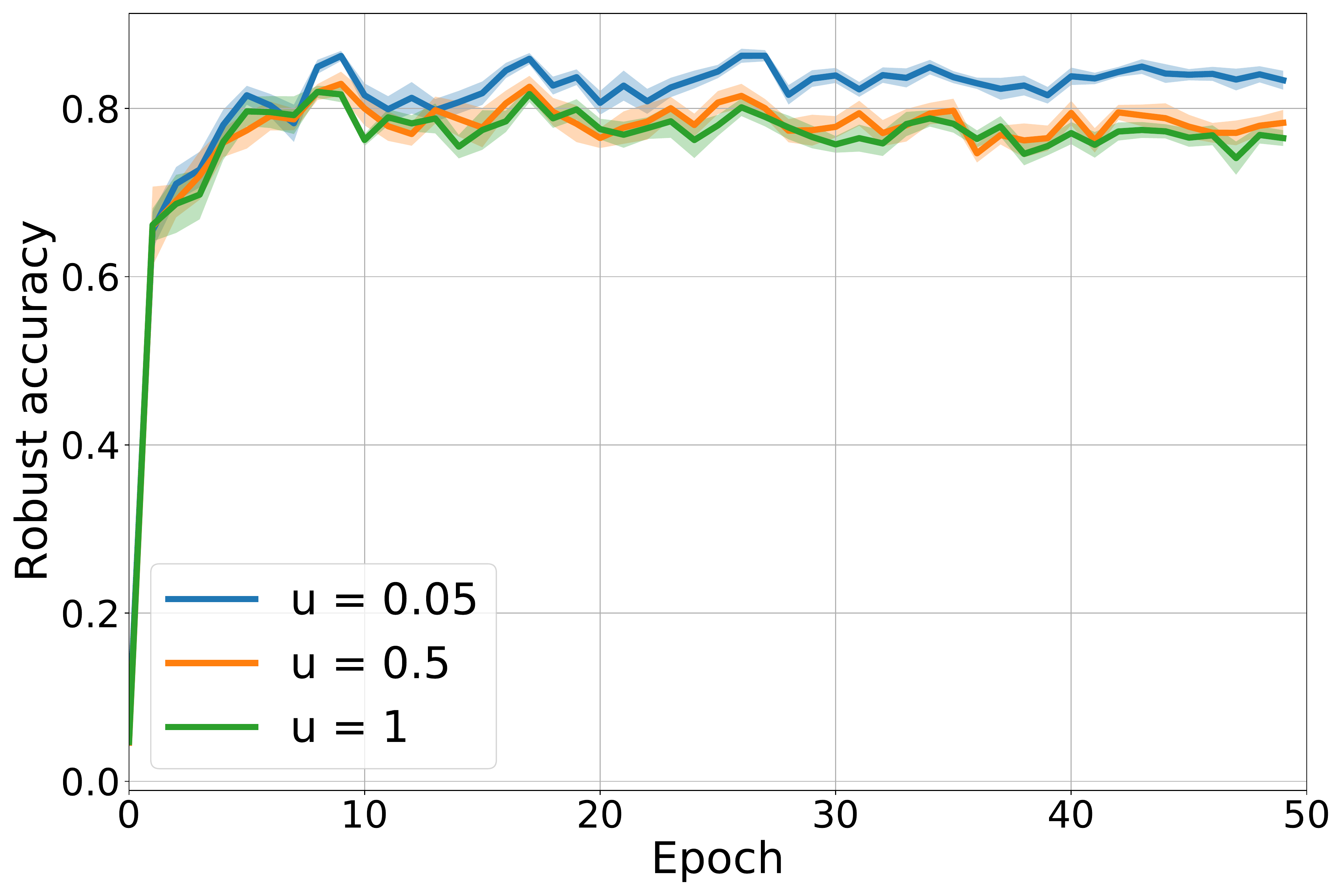}
    \caption{Robust accuracy during training of neural ODE model in the MNIST classification task. 
    Robust accuracy is computed with 9 different random seeds, the mean and the standard error is shown in the plot.
    Adversarial training is performed using FGSM random as described in \cite{wong2020fast}.}
    \label{fig:mnist_u}
    \end{subfigure}
    \caption{The value $u$ for the 2-nd order two stages Runge-Kutta method affects the robust accuracy of neural ODE model for MNIST classification task.}
    \label{fig::mnist_node_motivation}
\end{figure}

\begin{figure}
    \centering
    \includegraphics[width=0.49\linewidth]{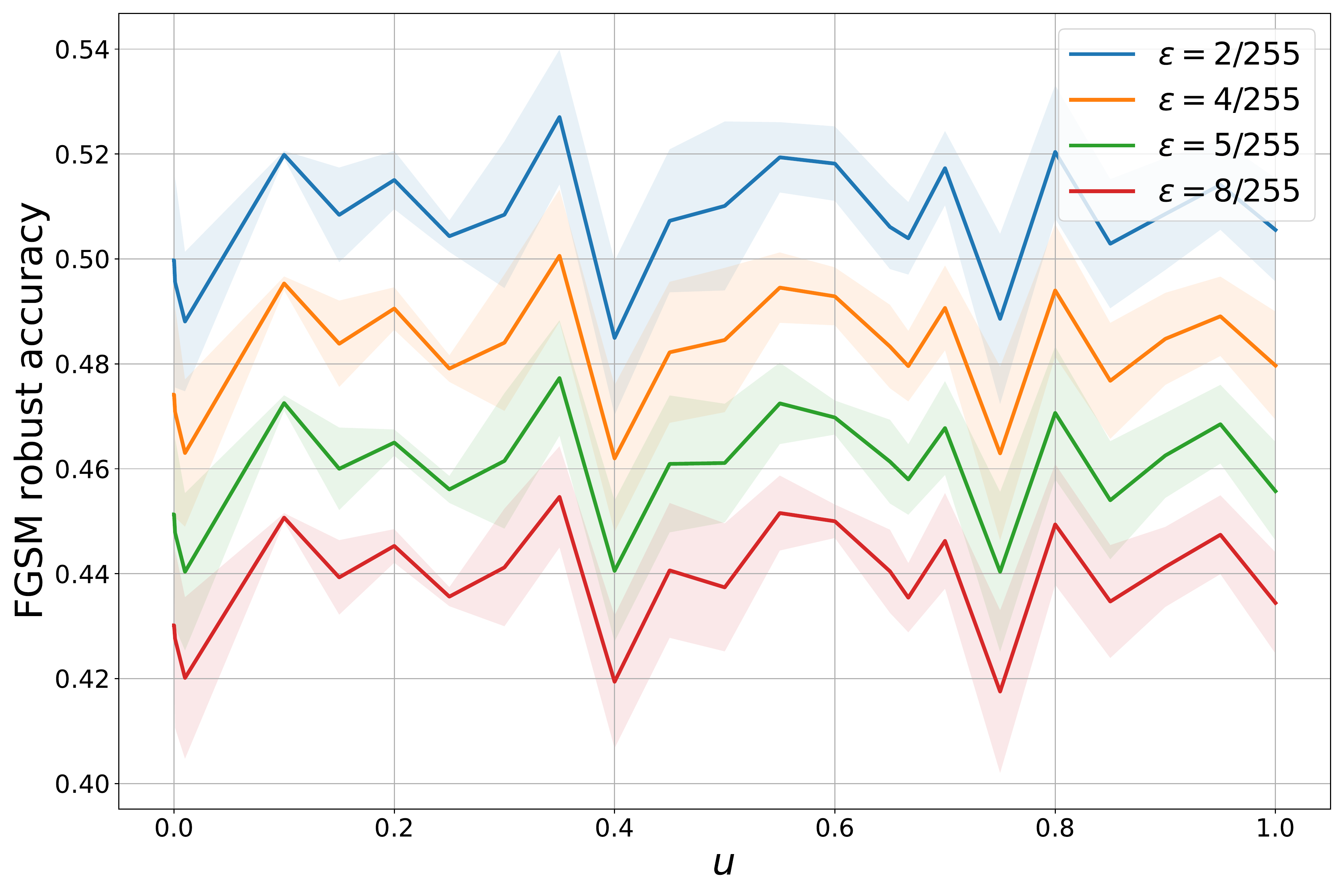}
    \caption{Comparison of robust accuracy corresponding to FGSM attack for range of $\varepsilon$, CIFAR-10 classification task.}
    \label{fig:cifar10_fgsm}
\end{figure}

\paragraph{CIFAR-10 classification task.}
We perform the same experiment on the CIFAR-10 classification task, but use FGSM attack and measure the robust accuracy for range of $\varepsilon$.
The obtained dependence of robust accuracy on the parameter of RK2 method is presented in Figure~\ref{fig:cifar10_fgsm}.
We observe that the maximum robust accuracy for every $\varepsilon$ is attained in the same values of $u$.
Details of this experiment are also given in Appendix.

In the next section, we provide approaches to adjust ODE solver parameter, implement these procedures and analyze them from the parametrizations properties perspective. 

%% file: 4_meta_node.tex
\section{Meta Neural ODE}
\label{sec::approach}
\textbf{Key idea.}
Solver parameters are modified during the neural ODE training by sampling from a given distribution. 
Hence, the model is trained using a large set of Runge-Kutta solvers instead of a single one, and yields better robustness to adversarial attacks without time overhead.

Since we want to adjust an ODE solver during training to improve neural ODE, the natural idea is to compute gradient of the loss function with respect to solver parameters and update them according to the gradient method altogether with weights in other layers.
We have tested this approach and figured out that the training is quite
unstable since the feasible parameters are not arbitrary and their clamping does not lead to desired improvement.
Thus, in this paper we introduce gradient-free methods of updating Runge-Kutta solver parameters during training. 

\paragraph{Solver switching and smoothing.}
During the neural ODE training,  at each epoch we randomly choose a solver from a pre-defined set of solvers to perform propagation through the model. If the set of solvers is continuous, we call this strategy \emph{solver smoothing}, otherwise, we refer to it as \emph{solver switching}.
The latter approach requires a  pre-defined set of parameterizations, each of which corresponds to one solver.  
Considering $s$-stage Runge-Kutta methods of order $p \leq 4, p=s$, each parametrization corresponds to one or two scalar values.

The sampling can be done uniformly or according  to some prior fixed distribution.
The benefits of switching is that it does not lead to computational overhead comparing to the single solver while trying to make the model more robust to the choice of the solver. However, such regime might make a neural ODE training via backpropogation difficult, if we have a limited number of solvers that exhibit different dynamics. That leads as to the smoothing approach, which can be considered as a continuous case of switching.
Smoothing regime requires to set in advance a parameterization of one initial solver. During the training, parameters for the next solver are sampled from a continuous distribution, whose mean corresponds with  parameters of the initial solver.
We expect that this approach leads to \emph{smoothing} of the trained dynamics and make it more robust to adversarial attacks.

\paragraph{Ensemble of models for free.}
When training a neural ODE using smoothing regime, we end up with a model, which performs well on a given task for a family of solvers. Hence, we can use this fact to build an ensemble of models to further improve the performance.

\section{Discussion}
The presented approach has two key ingredients.
The first one is the choice of values to initialize parameters.
We use the values that minimize the residual term in difference between ground-truth and approximate dynamics arising in definition~\ref{def::order} of the method order.
For example, a 2-nd order Runge-Kutta method with two stages has the residual term such that the value $u = \frac23$ minimizes it. 
More details on the derivation of such residual terms see in~\cite{wanner1996solving}.
However this approach does not lead to increasing of the robust accuracy.
The possible explanation of this observation is that the minimization of the residual term is less important for neural ODE model adversarial robustness.

The second ingredient is the distribution of the random variable that is used to generate new Runge-Kutta solver in every epoch.
The choice of this ingredient crucially depends on how coefficients in Butcher tableau relate to solver parameters.
The desired behaviour is that distribution of coefficients is unimodal.
Otherwise, the output of ODE block can be varied a lot which makes training harder.
For example, for a 2-nd order Runge-Kutta method with two stages the proper distribution is Cauchy distribution since it is preserved under addition and multiplication by scalar and inversion, see Figure~\ref{tab::sp2}.
Also, since it has heavy tails, we get higher variability in the used ODE solvers of the same order and number of stages. 
Note that, this method does not increase complexity of training since only sampling random variable is necessary.

\paragraph{Ensembling of solvers outputs.}
Another approach to adjust ODE solver during training neural ODE models is \emph{ensembling of solvers outputs}.
The idea of this method is to set ODE solvers corresponding to the same parameterization but from different values of parameters, compute trajectories with these solvers and average the computed trajectories with some pre-defined weights.
This method requires extra costs since multiple trajectories are computed.
However, the resulting dynamic approximation takes into account dynamics generated by different ODE solvers and therefore is more robust to the ODE solver choice.  
We tested this approach in CIFAR-10 classification task, but it provides insignificant improvement of robust accuracy.  

%% file: 5_experiments.tex
\section{Experiments}
To illustrate the approach presented in previous sections we perform numerical experiments with MNIST and CIFAR-10 classification tasks.
We test FGSM and PGD attacks and different values of $\varepsilon$.

\subsection{Solver parameterization does matter}
\begin{figure}[!h]
\begin{center}
    \includegraphics[width=0.5\linewidth]{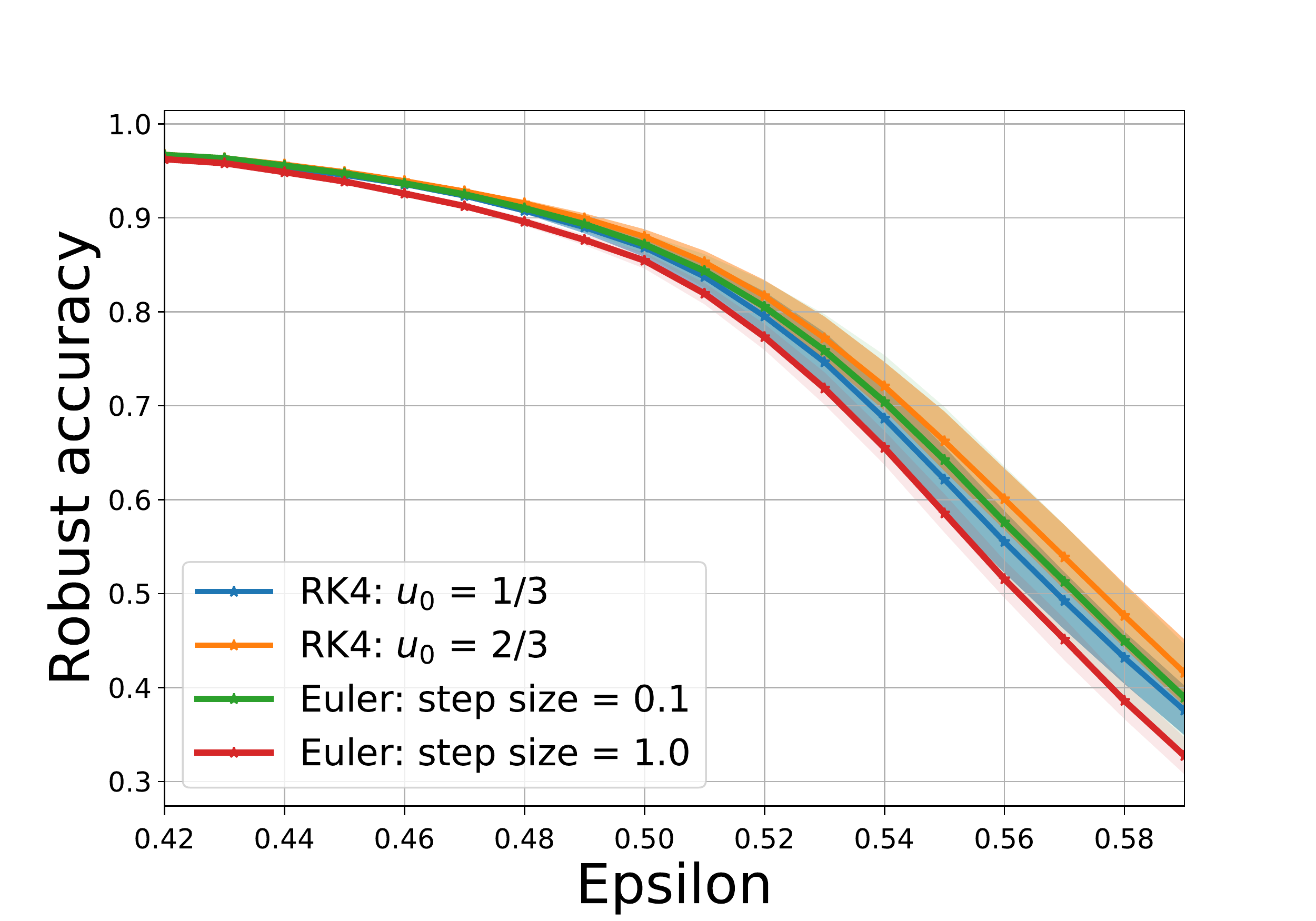}
\end{center}
\caption{Robust accuracy varies for different parameterizations and step sizes.}
\label{fig:robust_acc_together}
\end{figure}

In this section we provide experiments that motivate to explore the influence of solver parameterizations on neural ODEs performance. 
In \cite{hanshu2019robustness} they pointed out that neural ODE models are more robust than CNN models for some computer vision tasks. 
Inspired by this paper, we decided to move further and consider the influence of different ODE solvers to the robust accuracy of the model. 
In Figure~\ref{fig:robust_acc_together} we plot the robust accuracy for several values of $\varepsilon$ for the model trained on MNIST dataset. 
We demonstrate that 4-stage Runge-Kutta methods of 4-th order with different parameterizations yield  different level of robustness. 
We depicted at the same image 10-step Euler solver with a step size equals to 0.1, as it has been used in~\cite{hanshu2019robustness} and 1-step Euler for the comparison.  


\subsection{Solver smoothing to improve robustness}

In this section we demonstrate the effect of applying solver smoothing on CIFAR-10 classification task. 
We choose an architecture of the following type: (Conv layer $\to$ PreResNet block $\to$ ODE block $\to$ PreResNet block $\to$ ODE block $\to$ GeLU $\to$ AveragePooling $\to$ FullyConnected Layer). 
We use GeLU as activation functions inside ODE and PreResNet blocks. 
We chose an 8-step 2-stage Runge-Kutta solver with $u=0.5 $ as an initial solver for our solver smoothing strategy. 
We trained the model using SGD with momentum $0.9$ and a Cyclic LR schedule (one cycle per all iterations following the technique in~\cite{wong2020fast})  
To measure robust accuracy, we run FGSM attacks with $\epsilon=8/255$.
Also, during our experiments we have observed that solver smoothing allows to train models with higher learning rates.  
Table~\ref{tab:standard_smoothing} provides robust accuracies for 8-step 2-stage Runge-Kutta solver with $u=0.5$ if the solver smoothing is applied. 
During the training we sampled solver at each epoch from the normal distribution with zero mean and $\sigma = 0.0125$.
We observe the better robust accuracy if solver smoothing is applied, especially in the case of PGD attack.


\begin{table}[!ht]
    \centering
    \caption{Comparison of robust accuracy for different training schedule and attacks tested on the CIFAR-10 classification task, $\epsilon=2/255$. Solver smoothing setting: normal distribution, $\sigma = 0.0125$. Mean and standard error are computed across 3 random seeds.}
    \begin{tabular}{ccc}
    \toprule
         Training schedule & FGSM & PGD\\
        \midrule
 Standard & 33.02 $\pm$ 1.08  & 31.81 $\pm$ 1.03\\
 Solver smoothing &  33.64 $\pm$ 0.73 &  32.78 $\pm$ 0.65\\
 \bottomrule
    \end{tabular}
    \label{tab:standard_smoothing}
\end{table}

\subsection{Solver smoothing on top of adversarial training}
We also show that the adversarial training combined with the solver smoothing technique leads to more robust models, see Table~\ref{tab:adv_smooth}.
Worth noting that this is done without computational overhead.
Further improvement can be obtained by varying the hyper-parameters of the solver smoothing procedure.

\begin{table}[!ht]
    \centering
    \caption{Comparison of robust accuracy for different training schedule and attacks tested on the CIFAR-10 classification task, $\epsilon=8/255$. Solver smoothing setting: normal distribution, $\sigma = 0.0125$. Mean and standard error are computed across 3 random seeds.}
    \begin{tabular}{p{3cm}cc}
    \toprule
         Training schedule & FGSM & PGD \\
        \midrule
 Adversarial training & 40.86 $\pm$ 0.26 & 36.13 $\pm$ 0.13\\
 Solver smoothing \& adversarial training & 41.39 $\pm$ 0.01  &  36.37 $\pm$ 0.15\\
 \bottomrule
    \end{tabular}
    \label{tab:adv_smooth}
\end{table}


%% file: 6_conclusion.tex
\section{Conclusion}
In this study, we consider different parameterizations of the standard Runge-Kutta methods and study its influence on the neural ODE models training. 
We observe that test accuracy and robust accuracy after FGSM and PGD attacks depend on the used ODE solver even among the solvers from the same parametric family.
This observation leads to the idea of adjusting the ODE solver parameters to improve test accuracy and robustness of the trained neural ODE model. 
We propose the smoothing procedure that artificial noises ODE solver parameter in every epoch during training to make the output of ODE block more robust to perturbations of the trajectory.  
This procedure is tested in a standard benchmark and gives a more accurate and robust model than the baseline approach that uses the single ODE solver.
The presented approach can be extended to other parameterizations of ODE solvers to improve the robustness of the neural ODE models.

%% file: 7_appendix.tex
\section{Details about training neural ODE models}

In experiments, we focused on two benchmarks: MNIST and CIFAR-10 classification tasks.
Below we provide detailed description of models, optimizers and other hyperparameters for every task.

\subsection{MNIST}
\label{sec::mnist}
We perform two experiments on MNIST dataset to produce Figures 1 and 3 in the main text.
Both experiments were run with the same model, optimizer and hyperparameter values that are presented below. 
The model for the MNIST dataset is a sequentially stacked three neural networks.
The first one is a standard feedforward neural network of the following structure

\begin{enumerate}
    \item \texttt{Conv2d(1, 64, kernel\_size=(3, 3), stride=(1, 1))},
    \item \texttt{GroupNorm(32, 64, eps=1e-05, affine=True)},
    \item \texttt{ReLU(inplace=True)},
    \item \texttt{Conv2d(64, 64, kernel\_size=(4, 4), stride=(2, 2), padding=(1, 1))},
    \item \texttt{GroupNorm(32, 64, eps=1e-05, affine=True)},
    \item \texttt{ReLU(inplace=True)},
    \item \texttt{Conv2d(64, 64, kernel\_size=(4, 4), stride=(2, 2), padding=(1, 1))}.
\end{enumerate}
The second one is an ODE block with the following right-hand side $f(\vz, t)$ structure.

\begin{enumerate}
    \item \texttt{GroupNorm(32, 64, eps=1e-05, affine=True)},
    \item \texttt{ReLU(inplace=True)},
    \item \texttt{ConcatConv2d(65, 64, kernel\_size=(3, 3), stride=(1, 1), padding=(1, 1)},
    \item \texttt{GroupNorm(32, 64, eps=1e-05, affine=True)},
    \item \texttt{ConcatConv2d},
    \item \texttt{GroupNorm(32, 64, eps=1e-05, affine=True)},
\end{enumerate}
where \texttt{ConcatConv2d} is a convolution, applied to a batch with an additional channel; all elements of this channel are equal to time $t$.
And the final one is the following network

\begin{enumerate}
    \item \texttt{ReLU(inplace=True)},
    \item \texttt{AdaptiveAvgPool2d(output\_size=(1, 1))},
    \item \texttt{Flatten()},
    \item \texttt{Linear(in\_features=64, out\_features=10, bias=True)}.
\end{enumerate}

The backward pass is performed with the standard autodiff technique in both experiments.
The optimizer is RMSprop with a cyclic learning rate scheduler (\texttt{base\_lr 1e-5, max\_lr 1e-3, step\_size\_up 2000, step\_size\_down 2000, triangular2 mode}).
The whole training process lasted for 50 epochs with a batch size 128. 
To compute robust accuracy in both experiments we use PGD attack with $\varepsilon = 0.3$, learning rate $2/255$ and $7$ iterations.

\subsection{CIFAR-10}

To describe the structure of the network for CIFAR-10, let us first introduce a so-called \texttt{PreBasicBlock(input\_channels, output\_channels, skip\_connection\_layer=None}, which consists of

\begin{enumerate}
\item \texttt{GeLU(inplace=True)},
\item \texttt{Conv2d(input\_channels, output\_channels, kernel\_size=(3, 3), stride=(1, 1), padding=(1, 1), bias=False)},
\item \texttt{GeLU(inplace=True)},
\item \texttt{Conv2d(64, 64, kernel\_size=(3, 3), stride=(1, 1), padding=(1, 1), bias=False)},
\item and an optional skip-connection: a sum of the previous layer output and the input batch, propagated through the \texttt{skip\_connection\_layer}
\end{enumerate}

The whole CIFAR-10 network looks as follows

\begin{enumerate}
\item \texttt{Conv2d(3, 64, kernel\_size=(3, 3), stride=(1, 1), padding=(1, 1), bias=False)},
\item \texttt{PreBasicBlock(64, 64, skip\_connection\_layer=Identity())},
\item ODEBlock with the right-hand side \\ \texttt{PreBasicBlock(64, 64, skip\_connection\_layer=None)},
\item \texttt{PreBasicBlock(64, 128, skip\_connection\_layer=Conv2d(64, 128, kernel\_size=(3, 3), stride=(1, 1), padding=(1, 1), bias=False))},
\item ODEBlock with the right-hand side \\ \texttt{PreBasicBlock(128, 128, skip\_connection\_layer=None)},
\item \texttt{AdaptiveAvgPool2d(output\_size=(1, 1))},
\item \texttt{Flatten()},
\item \texttt{Linear(in\_features=128, out\_features=10, bias=True)}.
\end{enumerate}

This model is used in all tests with CIFAR-10 dataset.   
The optimizer is SGD with momentum=$0.9$ and a cyclic learning rate scheduler.
For Fig~2 and Table~2 from the main test we use (\texttt{base\_lr 1e-5, max\_lr 0.2, step\_size\_up 6240, step\_size\_down 6240, triangular2 mode}), the whole training process lasted for 64 epochs with a batch size 256.
For Table~1 and Table~2 in supplementary we use (\texttt{base\_lr 1e-7, max\_lr 0.1, step\_size\_up 3186, step\_size\_down 3186, triangular2 mode}), the whole training process lasted for 36 epochs with a batch size 256.

In our experiments we split CIFAR-10 dataset into traininig (40000), validation (10000) and test (10000) parts. We train models using training set, choose best model using validation set, and report final accuracy on the test set.

\section{Runge-Kutta methods of the 4-th order with 4 stages}
In this section we provide different parameterizations of Runge-Kutta methods of the 4-th order with 4 stages.
This class of Runge-Kutta methods induce the following system of equations on the Butcher tableau coefficients:
\[
\begin{cases}
    \sum_{i}b_i = 1,\\ 
    \sum_{i}b_ic_i = \frac{1}{2},\\ 
    \sum_{i}b_ic_i^{2} = \frac{1}{3},\quad
    \sum_{i, j}b_iw_{ij}c_j = \frac{1}{6},\\
    \sum_{i}b_ic_i^{3} = \frac{1}{4}, \quad
     \sum_{i, j}b_ic_iw_{ij}c_j = \frac{1}{8},\\ 
     \sum_{i, j}b_iw_{ij}c_j^{2} = \frac{1}{12}, \quad
     \sum_{i, j, k}b_iw_{ij}w_{jk}c_k = \frac{1}{24}.
\end{cases}
\]
This system is much more difficult compared with system corresponding to RK2 methods.
Therefore, it induces four parameterizations that we call $u_1, u_2, u_3$, and $uv$.
Parameterizations $u_1, u_2, u_3$ use the single parameter $u \neq 0$ and the resulting Butcher tableaux are presented in Figure~\ref{fig::butcher_tableau_u123}.
We highlight that parameterization $u_2$ gives the standard RK4 method for $u = \frac13$.
Thus, we can compare the standard RK4 method with other Runge-Kutta methods, which have the same order and number of stages, with respect to test accuracy and robust accuracy against PGD and FGSM attacks.
Also, based on the comparison results, we can adjust parameter $u$ to improve model quality.

\begin{figure}[!ht]
\centering
\begin{subfigure}[t]{0.45\textwidth}
\centering
\scalebox{1}{
\begin{tabular}{ c|cccc } 
$0$ & $0$ & &  &  \\ 
$\frac12$ & $\frac12$ & $0$ &  & \\ 
$0$ & $-\frac{1}{12u}$ & $\frac{1}{12u}$ & $0$ & \\ 
$1$ & $-\frac{1}{2} - 6u$ & $\frac{3}{2}$ & $6u$ & 0 \\ 
\hline
 & $\frac{1}{6} - u$ & $\frac23$ & $u$ &  $\frac{1}{6}$ \\ 
\end{tabular}
}
\caption{Butcher tableau for $u_1$ parameterization}
\end{subfigure}%
~
\begin{subfigure}[t]{0.45\textwidth}
\centering
\scalebox{1}{
\begin{tabular}{ c|cccc } 
$0$ & $0$ & &  &  \\ 
$\frac12$ & $0$ & $0$ &  &  \\ 
$\frac12$ & $\frac12 - \frac{1}{6u}$ & $\frac{1}{6u}$ & $0$ & \\
$1$ & $0$ & $1 - 3u$ &  $3u$ & $0$ \\ 
\hline
 & $\frac16$ & $\frac23 - u$ & $u$ & $\frac16$ \\
\end{tabular}
}
\caption{Butcher tableau for $u_2$ parameterization. 
Note that, if $u= \frac{1}{3}$, we get the standard Runge-Kutta method.}
\end{subfigure}
\\
\begin{subfigure}{0.45\textwidth}
\centering
\scalebox{1}{
\begin{tabular}{ c|ccccc } 
$0$ & $0$ & &  &  \\ 
$1$ & $1$ & $0$ &  &   \\ 
$\frac12$ & $\frac38$ & $\frac18$ & $0$ &   \\ 
$1$ & $1 - \frac{1}{4u}$ & $-\frac{1}{12u}$ &  $\frac{1}{3u}$ & $0$ \\ 
\hline
 & $\frac16$ & $\frac16 - u$ &  $\frac23$ & $u$ \\ 
\end{tabular}
}
\caption{Butcher tableau for $u_3$ parametrization}
\end{subfigure}
\caption{General form of Butcher tableaux corresponding to parameterizations $u_1, u_2, u_3$.}
\label{fig::butcher_tableau_u123}
\end{figure}

The parameterization $uv$ depends on two scalar values $u$ and $v$ such that $u \neq v, u \neq 0, 1, \frac12$ and $v \neq 0, 1$.
This parameterization induces the Butcher tableau presented in Table~\ref{tab::bucther_uv}.
Note that if $u = \frac{1}{3},\ v = \frac{2}{3}$, we obtain $3/8$ Runge-Kutta method~\cite{butcher1996history}.
The investigation of the optimal pair $(u, v)$ is more difficult than for parameterizations $u_1, u_2, u_3$, but the introduced solver smoothing procedure is still implementable without additional costs.

\begin{figure}[!ht]
    \centering
    \scalebox{1.}{
    \begin{tabular}{ c|cccc } 
$0$ & $0$ & &  &  \\ 
$u$ & $u$ & $0$ &  &   \\ 
$v$ & $v+\frac{u v-v^2}{2 u-4 u^2}$ & $\frac{(u-v) v}{2 u (-1+2 u)}$ & $0$ &   \\ 
$1$ & $\frac{2-5 v+4 v^2+4 u^2 \left(1-3 v+3 v^2\right)-3 u \left(2-5 v+4 v^2\right)}{2 u v (3-4 v+u (-4+6 v))}$ & $\frac{(-1+u) \left(-2+u+5 v-4 v^2\right)}{2
u (u-v) (3-4 v+u (-4+6 v))}$ & $\frac{(-1+u) (-1+2 u) (-1+v)}{(u-v) v (3-4 v+u (-4+6 v))}$ & $0$ \\
\hline
 & $\frac{1-2 u-2 v+6 u v}{12 u v}$ & $\frac{-1+2 v}{12 (-1+u) u (u-v)}$ & $\frac{1-2 u}{12 (u-v) (-1+v) v}$ & $\frac{3-4 v+u (-4+6 v)}{12 (-1+u) (-1+v)}$\\
\end{tabular}
}
\caption{The Butcher tableau for parameterization $uv$}
    \label{tab::bucther_uv}
\end{figure}
The parameterizations described above can be used to generate a Runge-Kutta method of the known order by setting arbitrary feasible values of corresponding parameters.